\documentstyle[aaainocpyrgt]{article}

\def\no{\hbox{\it not} }
\def\oor{{\,\hbox{\it or}\,}}
\def\iif{\leftarrow}

\def\citeyear{\cite}

\newtheorem{theorem}{Theorem}

\newtheorem{example}{Example}

\newtheorem{definition}{Definition}

\begin{document}
\title{A Splitting Set Theorem for Epistemic Specifications}
\author{Richard Watson\\
        Texas Tech University\\
        Department of Computer Science\\
        richard.watson@coe.ttu.edu}

\maketitle
\bibliographystyle{aaainocpyrgt}

\begin{abstract}
Over the past decade a considerable amount of research has been
done to expand logic programming languages to handle incomplete
information.  One such language is the language of epistemic
specifications. As is usual with logic programming languages, the
problem of answering queries is intractable in the general case.
For extended disjunctive logic programs, an idea that has proven
useful in simplifying the investigation of answer sets is the use of
splitting sets. In this paper we will present an extended
definition of splitting sets that will be applicable to epistemic
specifications.  Furthermore, an extension of the splitting set
theorem will be presented. Also, a characterization of stratified
epistemic specifications will be given in terms of splitting sets.
This characterization leads us to an algorithmic method of
computing world views of a subclass of epistemic logic programs.
\end{abstract}

\section{Introduction}
One of the most important areas in artificial intelligence is
knowledge representation.  Traditional logic programming has
proven itself to be a powerful tool for knowledge representation.
There are, however, limitations to the expressibility of
traditional logic programming. In an attempt to overcome some of
the these limitations, new logic programming formalisms were
introduced.  These new formalisms expand the traditional formalism
by including disjunction \cite{Min82}, classical negation
\cite{GL90}, or both (in the case of extended disjunctive logic
programs)\cite{GL91}.
These formalisms work well for certain classes of programs.
Unfortunately, these formalisms do not always allow for the
correct representation of incomplete information in the presence
of multiple belief sets.  As an attempt at solving this problem,
the language of epistemic specifications was introduced
\cite{Gel91,GP91}.  A good overview of each of the formalisms
mentioned above can be found in \cite{BG94}.

As is usual with logic programming languages, the problem of
answering queries is intractable in the general case.  It is often
useful, however, to find methods which simplify the query
answering task for certain subclasses of programs. In \cite{LT93}, 
the usefulness of splitting sets for the investigation of 
answer sets was shown. In this paper we will present an extended definition
of splitting sets that will be applicable to epistemic
specifications. This in turn leads to an extension of the
splitting set theorem from \cite{LT93}.  As with EDLPs, there is a
strong relationship between stratification and splitting sets.
Using these ideas, we can develop an algorithmic method
for computing world views of a subclass of epistemic logic
programs.

An overview of the syntax and semantics of epistemic specifications is
covered in Section 2.  In Section 3 we present splitting sets for epistemic
specifications and the main theorem of the paper, the splitting set theorem.
Finally, Section 4 contains a discussion of
stratification, how it is related to splitting sets, and an algorithm for
computing world views of stratified programs which is based the splitting set theorem.

\section{Epistemic Specifications}
The language of epistemic specifications is an extension of the
language of extended disjunctive logic programs (EDLPs).  In
addition to the normal operators in EDLPs, the language of
epistemic specifications also contains unary modal operators $K$
and $M$. $K$ should be read as {\em ``is known to be true''} while
$M$ is read as {\em ``may be believed to be true''}. Atoms are
defined in the usual way.  Literals in the language of epistemic
specifications are split into two types, {\em objective literals}
and {\em subjective literals}. An {\em objective literal} is
either an atom or an atom preceded by $\neg$ (classical negation).
A {\em subjective literal} is an objective literal preceded by
$K$, $M$, $\neg K$, or $\neg M$.  Given an objective literal, $L$,
we will refer to the set of four subjective literals that can be
built from $L$ as $SubLit(L)$.  Given a set of objective literals,
$U$, $SubLit(U) = \{X: L \in U$ and $X \in SubLit(L)\}$.

\subsection{Syntax}
The general form for rules in epistemic specifications is given in
\cite{Gel94}. In this paper we will restrict rules to the form:

\noindent
\begin{center}
$F_1 \oor\ \dots \oor\ F_n \iif G_1, \dots , G_k, \no\ G_{k+1}, \dots , \no\ G_m$
\end{center}

\noindent where $F_1, \dots , F_n$ are objective literals,  $G_1,
\dots , G_k$ are either objective or subjective literals, and
$G_{k+1}, \dots , G_m$ are objective literals.  This form differs
from the original only in the fact that in \cite{Gel94}, $G_{k+1},
\dots , G_m$ were also allowed to be subjective literals.  Notice
however that for any subjective literal, $G_i$, the value of $G_i$
can never be unknown and hence $\no\ G_i$ is always equivalent to
$\neg G_i$.  It can therefore easily be seen that the restricted
form of rules above can be used without any loss of
expressibility.

A collection of such rules will be referred to as an {\em
epistemic logic program} or an {\em epistemic specification}.
Given a rule, $r$,
\begin{itemize}
\item $head(r)$ refers to the the set of literals, $\{F_1, \dots , F_n\}$
which occur in the head of the rule.
\item $pos(r)$ refers to the set of all objective literals, $L$, such that
either
\begin{itemize}
\item $L = G_i$ for some $1 \leq i \leq k$, or
\item $G_i \in SubLit(L)$ for some $1 \leq i \leq k$.
\end{itemize}
\item $neg(r)$ refers to the set of literals, $\{G_{k+1}, \dots , G_m\}$.
\item $lit(r) = head(r) \cup pos(r) \cup neg(r)$.
\end{itemize}
Given a epistemic logic program, $\Pi$,
$Lit(\Pi)$ will denote the union of the sets $lit(r)$ for all $r
\in \Pi$. 

\subsection{Semantics}
We now move from the syntax of the language to the semantics. A
rule with variables is considered to be a shorthand for the set of
all ground instances of the rule. The truth or falsity of a
literal in an epistemic logic program is determined by the {\em
world views} of that program.  A {\em world view} is a collection
of sets of ground objective literals which satisfy certain
properties. An objective literal, $L$, is true with respect to a
collection of sets of literals, $W$, if it is true in each set in
that collection (i.e. for each set $A \in W$, $L \in A$). If $W$
is a collection of sets of objective literals and $L$ is an
objective literal then
\begin{itemize}
\item $KL$ is true with respect to $W$ (denoted
$W \models KL$) iff for each set $A \in W$, $L \in A$,
\item $W \models ML$ iff there exists an $A \in W$ such that $L \in A$,
\item $W \models \neg KL$ iff $W \not\models KL$, and
\item $W \models \neg ML$ iff $W \not\models ML$.
\end{itemize}
A literal is true with respect to an epistemic logic program if it
is true in every {\em world view} of that program. We will define
the concept of a {\em world view} of an epistemic logic program in
several steps.

First let us consider the case when $\Pi$ is an epistemic logic program which
does not contain $not$ and does not contain any subjective literals.  A set
of literals, $A$, is called a {\em belief set} of $\Pi$ iff $A$ is a minimal
set satisfying the following two conditions:
\begin{itemize}
\item For every rule $F_1 or \dots or F_n \leftarrow G_1, \dots, G_k \in \Pi$
if $G_1, \dots, G_k \in A$ then $\exists i, 1 \leq i \leq n$ such that $F_i \in A$,
\item If $A$ contains a pair of contrary literals then $A = Lit$.  (This belief set is called {\em inconsistent}.)
\end{itemize}

Next we consider an epistemic logic program, $\Pi$, which contains
$not$ but does not contain subjective literals (such programs are
extended disjunctive logic programs). For any such $\Pi$ and any
set $A \subset Lit(\Pi)$, let $\Pi^A$ be the program obtained from
$\Pi$ by deleting
\begin{itemize}
\item each rule that contains $not\ L$ in its body where $L \in A$, and
\item all formulas of the form $not\ L$ in the bodies of the remaining rules.
\end{itemize}
The resultant program $\Pi^A$ does not contain $not$ or subjective literals and
therefore its belief sets are as defined above.  We will say a set, $A$, of literals
is a belief set of $\Pi$ if $A$ is a belief set of $\Pi^A$.

Finally, let $\Pi$ be an arbitrary epistemic logic program.  Let, $W$, 
be any collection of sets of literals from $Lit(\Pi)$ and let $\Pi^W$ be the program obtained by
\begin{itemize}
\item removing each rule which contains a subjective literal, $L$, where $W \not\models L$, and
\item removing all subjective literals from the bodies of the remaining rules.
\end{itemize}
Notice that $\Pi^W$ does not contain subjective literals, therefore we can compute its
belief sets as previously described.  If $W$ is the set of all of the belief
sets of $\Pi^W$ then $W$ is a world view of $\Pi$.

We will say that a world view of an epistemic logic program is {\em consistent}
if it does not contain a belief set consisting of all literals.  We will
say an epistemic logic program is {\em consistent} if it has at least one
consistent non-empty world view.

Intuitively, a belief set is a set of literals that a rational
agent may believe to be true.  A world view is a set of belief
sets that a rational agent may believe to be true with respect to
that ``world''.

The following give examples of epistemic logic programs and their
world views. 

\begin{example}
Let $\Pi_1$ be the program which consists of the
rules:
\begin{enumerate}
\item $p(a)\ or\ p(b) \leftarrow$
\item $p(c) \leftarrow$
\item $q(d) \leftarrow$
\item $\neg p(X) \leftarrow \neg \ Mp(X)$
\end{enumerate}
The set
 $$W=\{\{q(d),p(a),p(c),\neg p(d)\},\{q(d),p(b),p(c),\neg
p(d)\}\}$$ consisting of two belief sets, can be shown to be the
only world view of $\Pi_1$.
\end{example}

\begin{example}
For the next example, consider the program, $\Pi_2$, consisting of
the following two rules:
\begin{enumerate}
\item $p(a) \leftarrow \neg Mq(a)$
\item $q(a) \leftarrow \neg Mp(a)$
\end{enumerate}
It can be seen that $\Pi_2$ has two world views: $W_1 = \{\{q(a)\}\}$
and $W_2 = \{\{p(a)\}\}$.
\end{example}

\begin{example}
As a final example, consider the program, $\Pi_3$ consisting of only
one rule, 
$$p(a) \leftarrow \neg Kp(a).$$
It can be shown that this
program does not have a world view.
\end{example}

In general, to find the world view of a epistemic logic program
one must either try all possible collections of sets of literals
or guess.  
It is infeasible to try all combinations since, even for the case
where the number of ground literals, $n$, is finite, there are $2^{2^n}$
possibilities. 
A guess-and-check method could possibly be used to find world
views but the problem is how to create an algorithm which
would make good ``educated'' guesses and would know when and if it has
found all the of the world views.

In this paper, we are primarily interested in presenting a means of
computing world views.  As a first step in achieving this goal, we
will limit ourselves to programs which have at most a finite number
of world views.  For the remainder of this paper we will only consider
epistemic logic programs which do not contain function symbols and 
have a finite number of constants and predicate symbols.  

\section{Splitting Sets}

In this section we will present a definition of splitting sets of
epistemic logic programs.  The definition is an extension of the
definition in \cite{LT93}.  We will also present a version of the
splitting set theorem that is applicable to epistemic logic
programs.

\begin{definition}[Splitting Set]
A set, $U$, of objective literals is a {\em splitting set} of a epistemic logic program,
$\Pi$, iff
\begin{itemize}
\item for every rule $r \in \Pi$, if $head(r) \cap U \not= 0$ then $lit(r) \subset U$, and,
\item if $\Pi$ contains $K$ or $M$, then for any objective literal, $p \in lit(\Pi)$, if $p \in U$
then $\overline{p} \in U$.
\end{itemize}
\end{definition}
If $U$ is a splitting set of $\Pi$, we also say that $U$ splits
$\Pi$.  The set of all rules $r \in \Pi$ such that $lit(r)
\subset U$ is denoted by $b_U(\Pi)$ and is called the {\em
bottom} of $\Pi$ with respect to $U$.  The set $\Pi \backslash
b_U(\Pi)$ is called the {\em top} of $\Pi$ with respect to $U$.

Using a splitting set, one can break the computation of a world
view of an epistemic specification into two parts, a bottom and a
top. The basic idea is to first compute the world view of the
bottom of the program.  The world view of the top can then be
computed, taking into consideration the what was already computed
for the bottom. Finally the two parts are merged together to get
the world view of the complete program.

The world view of the bottom can be computed without regard to the
top since no literal which occurs in the head of a rule of the top
can occur anywhere in the bottom. When computing the world view
for the top however, one needs to take the world view of the
bottom into consideration.
The world view of the bottom of the program can be used to
``reduce'' the top of the program.  We can
remove from the top those rules which cannot be satisfied because
the value of a literal computed in the bottom makes their bodies
false. From the remaining rules one can remove the portions of the
bodies of the rules that were determined to be true. The reduction
is performed in two steps; one for subjective literals and one for
objective ones.

To remove subjective literals we will introduce the idea of a {\em
restricted reduct}. 

\begin{definition}[Restricted Reduct]
Let $\Pi$ be an\\ epistemic logic program, $W$
be a collection of sets of literals, and $U$ be a set of literals.
The {\em restricted reduct} is the program obtained from $\Pi$ by:
\begin{enumerate}
\item removing from $\Pi$ all rules containing subjective formulae $G$ where
$G \in SubLit(U)$ and $W \not \models G$.
\item removing all other occurrences of subjective formula $G$ where $G \in SubLit(U)$.
\end{enumerate}
\noindent The resultant program will be denoted by $\Pi^{r(U,W)}$
and be referred to as the {\em reduct} of $\Pi$ with respect to
$W$, {\em restricted} by $U$.
\end{definition}
In our intended use, $\Pi$, would
be the top of a program, $U$, would be the set used to split the
program, and $W$ would be the world view of the bottom.

The following is an example of a restricted reduct.
\begin{example}
{\ }
\begin{tabbing}
MMM \= MMMMM \= \kill
Let \> $W = $ \> $\{\{a, \neg b,\ d \}, \{a, \neg d \}\}$, \+ \\
    $U = $ \> $\{a, \neg a,\ b, \neg b,\ c, \neg c \}$, and \\
    $\Pi =$ \> $ e \leftarrow a,\ M \neg b,\ f$ \+ \\
          $g \leftarrow Ka,\ h$ \\
          $i \leftarrow Mc$ \\
          $j \leftarrow Kd,\ k$ \- \- \\
then \> $\Pi^{r(U,W)} = $ \> $e \leftarrow a,\ f$ \+ \+ \\
     $g \leftarrow h$
\end{tabbing}
\end{example}

Next we consider objective literals.
Recall that the world view of the bottom of a program is in
essence a set of belief sets, all of which are different.  Because
of this, the truth or falsity of the objective literals in the
bodies of rules of the top may vary with respect to each belief
set.  Due to this fact, after performing the reduction described
below, rather than being left with a single program, we have, in
general, a different partially evaluated top for each belief set
of the bottom.
\begin{definition}[Partial Evaluation]
Given two sets of objective literals, $U$ and $X$, and an
epistemic specification, $\Pi$, for which none of the literals
from $U$ or $X$ occur subjectively in its rules, then $e_U(\Pi,X)
= \{ r'$: $\exists$ rule $r \in \Pi$ such that $pos(r) \cap U
\subset X$ and $neg(r) \cap U$ is disjoint from $X$, $r'$ is the
rule which results from removing each sub-formula of the form $L$
or $\no\ L$ from $r$, where $L \in U \}$.  We refer to
$e_U(\Pi,X)$ as the {\em partial evaluation} of $\Pi$ with respect
to $X$.
\end{definition}
 Here again, in our intended use $\Pi$ would be the top of
the program, $U$ would be the splitting set used, and $X$ would be
one of belief sets from the world view of the bottom.

As was mentioned above, after taking the restricted reduct of the
top and then finding the partial evaluation of the result with respect
to each of the belief sets of the bottom, we are often left with
multiple ``tops''. We cannot simply take the world view of each
``top'' and merge them together. The reason for this is that it
does not guarantee that the truth of subjective literals in the
merged world view are the same as they were in each ``top''. To
handle this problem we introduce the idea of a {\em multi-view}.

\begin{definition}[Multi-view]
Given epistemic logic\\
 programs $\Pi_1, \dots ,\Pi_n$, then a collection of sets of
objective literals, $W$,
is a {\em multi-view} of $\Pi_1, \dots ,\Pi_n$
iff
\begin{enumerate}
\item  $W = (\bigcup_{i=1}^{n} ans(\Pi_i^W)) \backslash \{Lit\}$ ( if
$\exists i$ s.t. $ans(\Pi_i^W)$ is consistent)
\item  $W = \{\{Lit\}\}$ (otherwise)
\end{enumerate}
   A multi-view, $W$, is {\em consistent} iff
$W \neq \{\{Lit\}\}$. For each $\Pi_i$, the set of all belief sets of
$\Pi_i^W$ is called
the {\em restricted view of $\Pi_i$ with respect to W}.
\end{definition}

Here a simple example of a multi-view.
\begin{example}
{\ }
\begin{tabbing}
MMMMM \= MMM \= MMM \= \kill \> \+ If \> \+ $\Pi_1 = $ \> \+ $a
\leftarrow $\\ $b \leftarrow$\\ $c \leftarrow Kb $ \- \- \- \\ \>
\+ and \> \+ $\Pi_2 = $ \> \+ $a \leftarrow $\\ $c \leftarrow Kb $
\- \-
\end{tabbing}
then $\Pi_1, \Pi_2$ has only one multi-view, \{\{a, b\},\{a\}\}.
\end{example}

Before we present the main theorem of the paper we must first present a new notation 
and a definition.
Given a collection of sets of objective literals, $W$, and a set
of literals, $U$, then $$W|_U = \{X :  \exists W_i \in W, X = W_i \cap U\}.$$

\begin{definition}[Safe]
Given an Epistemic Specification $\Pi$ with splitting set $U$ such
that $\Pi_U = b_U(\Pi)$ and $\Pi_{\overline{U}} = \Pi \backslash
\Pi_U$, $\Pi$ is said to be {\em safe with respect to U} iff
$\forall W = \{W_1, \dots , W_n \}$ if $\{ W_1|_U, \dots , W_n|_U
\} \subseteq ans(\Pi_U^W)$ then $\forall A \in ans(\Pi_U^W)
:(e_U(\Pi_{\overline{U}}^W,A))$
 is consistent.
\end{definition}

\begin{theorem}
{\rm
Let $\Pi$ be an epistemic specification, $U$ be a splitting set of $\Pi$ such
that $\Pi$ is safe with respect to $U$. If we denote $b_U(\Pi)$ as $\Pi_U$, and
$\Pi \backslash \Pi_U$ as $\Pi_{\overline{U}}$ then:
\begin{enumerate}

\item If $$X = \{X_1, \dots , X_n\}$$ is a consistent world view of $\Pi_U$ and
$Y$ is a consistent multi-view of
$$(e_U(\Pi_{\overline{U}}^{r(U,X)}, X_1), \dots ,
e_U(\Pi_{\overline{U}}^{r(U,X)}, X_n))$$ then if $W = \{W_i : W_i =
X_j \cup Y_k$, where $X_j \in X, Y_k \in
ans((e_U(\Pi_{\overline{U}}^{r(U,X)}, X_j))^Y)$ and $X_j \cup Y_k$
is consistent $\} \neq \{ \}$ then $W$ is a consistent world view
of $\Pi$.

\item If $W$ is a consistent world view of $\Pi$ then $\exists X,Y$
such that $X$ is a world view of $\Pi_U$, $Y$ is a multi-view of
$$(e_U(\Pi_{\overline{U}}^{r(U,X)}, X_1), \dots ,
e_U(\Pi_{\overline{U}}^{r(U,X)}, X_n))$$ and $$\forall W_i \in W (
W_i|_U \in X$$ and $$W_i|_{\overline{U}} \in
ans((e_U(\Pi_{\overline{U}}^{r(U,X)}, W_i|_U))^Y)$$

\end{enumerate}
}
\end{theorem}

In the above theorem we require that the splitting set be safe
with respect to the program.  As we will show, this restriction is
important. If one or more or the belief sets of the bottom does
not have a consistent extension to the top, the value of
subjective literals defined in the bottom may change.  In this
case, the above method may not compute a correct world view.  

\begin{example}
Consider the program, $\Pi_4$, with the following rules:
\begin{enumerate}
\item $p(a)\ or\ p(b) \leftarrow$
\item $p(c) \leftarrow Mp(b)$
\item $p(d) \leftarrow p(b)$
\item $\neg p(d) \leftarrow p(b)$
\end{enumerate}
If we split the program using $$U = \{p(a),\neg p(a), p(b),\neg
p(b), p(c), \neg p(c)\}$$ as a splitting set, then $b_U(\Pi_4)$,
which consists of rules 1 and 2, has one world view which contains
2 belief sets, $\{p(a),p(c)\}$ and $\{p(b),p(c)\}$.  With respect
to the belief set $\{p(b),p(c)\}$, however, the top of the program
is inconsistent.  Using the method from the theorem above, not
requiring the program be safe, we get one ``world view'':
$\{\{p(a), p(c)\}\}$. It can easily be seen however, that this is
not a world view of $\Pi_4$. The only world view of the program is
$\{\{p(a)\}\}$.
\end{example}

 The error occurred because, since $p(b)$ was
``possible'' in the world view the bottom we concluded $p(c)$ was
therefore true even though we later find that $p(b)$ is no longer
``possible'' after the computation of the top.

As can be seen from the definition, determining if a splitting set
of a program is safe may be as difficult as finding the world
views. We will give a property which is more intuitive
and easier to check.  While it is less general, it is reasonable
and encompasses a large number of interesting programs.
Before we present the condition, we must first define {\em satisfies}.

\begin{definition}[Satisfies]
Given a program $\Pi$ and a collection of sets of literals from $Lit(\Pi)$, denoted $W$,
then we will say $W$ {\em satisfies} the body of a rule, $r \in \Pi$, if each literal
in the body is true with respect to $W$.  We say $W$ satisfies $r$ if either $W$ does
not satisfy the body of $r$ or at least one literal in the head of $r$ is true with respect
to $W$.
\end{definition}

We now present the property.

\begin{definition}[Guarded]
We will say that a program $\Pi$ is {\em guarded} with respect to
a splitting set $U$ if
\begin{itemize}
\item $\Pi$ does not contain subjective literals, or
\item for every pair of rules $R_1, R_2 \in \Pi \backslash b_U(\Pi)$ and for every collection of sets of
literals from $Lit(\Pi)$, denoted as $W$, if $head(R_1)$ and
$head(R_2)$ contain contrary literals and $W$ satisfies all of the rules in $b_U(\Pi)$ then
either $W$ does not satisfy the body of $R_1$ or $W$ does not satisfy
the body of $R_2$.  Note that a rule with an empty head can be rewritten as a rule
which has the predicate $\neg true$ as the head and by adding the rule $$true \leftarrow$$
to the program.  A program containing rules with empty heads is guarded with respect to $U$ 
if the program rewritten without such rules is.
\end{itemize}
\end{definition}

It can be shown that, given any program $\Pi$ with splitting set
$U$, if $\Pi$ is guarded with respect to $U$ then $U$ is safe with
respect to $\Pi$.

\section{Splitting and Stratification}

In this section we will give a definition of stratification for epistemic
logic programs, show how it relates to splitting sets, and illustrate
how the splitting set theorem can be used to simplify the computation of the
world view of a stratified epistemic logic program.  We will start out
with the definition of stratification.

\begin{definition}[Stratification]
A partitioning $$\pi_0,\ldots ,\pi_z$$ of the set of all literals
 of an epistemic logic program, $\Pi$, is a {\em
stratification} of $\Pi$, if for any literal, $L_1 \in \pi_i$, then
$$\neg L_1 \in \pi_i$$
 and for any other literal $L_2$ in $Lit(\Pi)$ and any rule $r \in \Pi$:
\begin{itemize}
\item if $L_1, L_2 \in head(r)$ then $L_2 \in \pi_i$.
\item if $L_1 \in head(r)$ and $L_2$ occurs objectively in $pos(r)$ then
there exists an $j \leq i$ such that $L_2 \in \pi_j$.
\item if $L_1 \in head(r)$ and $L_2 \in neg(r)$ or $L_2$ occurs subjectively
in $r$, then there exists $j < i$ such that $L_2 \in \pi_j$. 
\end{itemize}

This stratification of the literals defines a stratification of the
rules of $\Pi$
to strata $\Pi_0, \dots , \Pi_k$ where a strata $\Pi_i$ contains all of the
rules of $\Pi$ whose heads consists of literal from $\pi_i$.
A program is called {\em stratified} if it has a stratification.
\end{definition}

It can easily be seen that, given a stratified epistemic logic program, $\Pi$, with
stratification $\pi_0,\ldots ,\pi_z$, the set of literal $U_i$ such that
$$U_i = \bigcup_{j=1}^i \pi_j$$ is a splitting set of $\Pi$.  
With each stratified epistemic logic program we will then associate a sequence 
$U_0, \dots, U_z$ of splitting sets formed as described.

This leads us to an algorithm for computing the world view a safe,
stratified epistemic specification. Given an epistemic
specification, $\Pi$, with stratification $\pi_0,\ldots ,\pi_n$
and associated splitting sets $U_0, \dots , U_n$, such that $\Pi$
is safe with respect to $\{\ \}$ and each $U_i$, we can compute
the world view of $\Pi$ as follows:
\begin{enumerate}
\item Using the splitting set theorem, compute the world view, $W_1$, of
$\Pi_0 \cup \Pi_1$ with splitting set $U_0$.  Note that $b_{U_0} =
\Pi_0$ and, by the definition of stratification, it does not
contain $not$ or any subjective literals. $\Pi_0$ is also safe
with respect to $\{\ \}$.  From these two facts, it can be seen
that $\Pi_0$ has a unique, consistent, world view which consists
of all the belief sets of the EDLP $\Pi_0$.
\item Given the world view, $W_{i-1}$, of $\Pi_0 \cup \dots \cup \Pi_{i-1}$,
the world view, $W_i$, of $\Pi_0 \cup \dots \cup \Pi_i$ can be
computed using the splitting set theorem with the splitting set $U_{i-1}$.
\end{enumerate}
Notice that $W_n$ is the world view of $\Pi$. It can be seen from
the definition of stratification that, in each step of the
algorithm above, when we take the restricted reduct of the top of
program we are left with a program which does not contain
subjective literals.  The multi-view is therefore simply the union
of the world views obtained by taking the restricted reduct of the
top and partially evaluating with respect to one of the belief
sets of the world view of the bottom. To compute the world view of
a safe, stratified, epistemic logic program therefore, one only
needs to be able to compute the belief sets of extended
disjunctive logic programs.

The following theorem, which is a slightly modified version of a
theorem from \cite{Wat94}, also follows from the results above.

\begin{theorem}
{\rm Given any stratified, epistemic logic program, $\Pi$, which
is safe with respect to $\{\ \}$ as well as each of the splitting
sets associated with its stratification, the program $\Pi$ has a
unique, consistent, world view. }
\end{theorem}

\section{Conclusion}
In this paper, we expanded the results from \cite{LT93} to include
epistemic logic programs.  We also presented definitions of what
it means for a epistemic logic program to be {\em safe}, {\em
guarded}, and {\em stratified}.  This led to an algorithmic method
for computing world views of a subclass of epistemic logic
programs.

It should be noted that the belief sets of an extended disjunctive
logic program are simply the answer sets \cite{GL91} of that
program.  Recently, there have been considerable advances in the
computation of such answer sets.  One such system which shows
great promise is DLV \cite{Leo97}.  Using their system and the
results in this paper, it should be a reasonable task to create a
inference engine for the subclass of epistemic logic programs
mentioned here.

As this paper is meant to form a basis for the computation of 
world views, we restricted ourselves to epistemic logic programs 
with a finite number of finite world views.  We believe that the
theorem presented here can be expanded to cover programs with an
infinite number of infinite world views.

\section{Acknowledgements}
The author would like to thank Michael Gelfond and the anonymous 
reviewers for their helpful comments.

\bibliographystyle{aaainocpyrgt}

\end{document}